\pdfoutput=1
\PassOptionsToPackage{backref=page,breaklinks=true}{hyperref}
\documentclass[11pt]{article}

\newif\ifauthordecided
\authordecidedtrue

\newif\ifarxiv
\newif\ifcameraready

\arxivtrue         % true = arXiv submission version (use preprint)
\camerareadyfalse  % true = accepted and camera-ready

\newif\ifperfect
\perfectfalse

\usepackage{times}
\usepackage{latexsym}
\usepackage{natbib}
\usepackage{tabularx}
\usepackage{array}
\usepackage{xcolor}
\usepackage{float}
\usepackage[preprint]{acl}
\usepackage{capt-of}

\usepackage[most]{tcolorbox}
\definecolor{CB_babyblue}{HTML}{A4DDED}
\newcommand{\fon}[1]{\fontfamily{#1}\selectfont}
\tcbset{prompt/.style={
    enhanced,
    size=fbox,
    boxrule=3pt,
    arc=2mm,
    auto outer arc,
    left=10pt,
    right=10pt,
    top=10pt,
    bottom=10pt,
    colback=CB_babyblue!15,
    colframe=CB_babyblue!30,
    coltitle=CB_babyblue!25!black, 
    fontupper=\fon{cmtt}, 
}}

\usepackage[T1]{fontenc}
\usepackage[utf8]{inputenc}

\usepackage{microtype}

\usepackage{inconsolata}

\usepackage{booktabs}
\usepackage{graphicx}
\usepackage{xspace}
\usepackage{url}

\title{From UNDRR Reports to Event Records: Schema-Constrained LLM Extraction of Georeferenced Disasters}

\ifauthordecided
\author{Camilla Andreozzi%
\thanks{Research done while the author was an intern at the United Nations Office for Disaster Risk Reduction (UNDRR).}
\\
  ETH Zürich \\
  \texttt{candreozzi@ethz.ch} \\\And
  Phuong-Anh Nguyen-Le%
  \\
  University of Maryland \\
  \texttt{nlpa@umd.edu} \\\AND
  Zhijing Jin \\
  MPI \& University of Toronto\\
  \texttt{zjin@cs.toronto.edu} \\\And
  Revati Mani\\
  United Nations Office for Disaster Risk Reduction\\
  \texttt{revati.mani@un.org}
}
\fi

\begin{document}

\maketitle
\begin{abstract}
Disaster-risk-reduction archives describe hazard events in prose that databases such as EM-DAT \citep{delforge2025emdat} cannot ingest directly. We present an LLM pipeline that generates candidate georeferenced event records using a controlled hazard vocabulary and fixed schema, retaining evidence for review. Applied to 10{,}000 documents from PreventionWeb, the knowledge hub managed by UNDRR, it produced 3{,}572 records from 1{,}913 documents across 24 hazard types and resolved 81\% of location mentions to OpenStreetMap geometries. On 171 human-positive document windows from a stratified 217-document reference set, GPT-5 achieved 86.0\% pooled attribute F1, versus 44.2\% for the spaCy--gazetteer baseline. Evaluation pools hazard families, location strings, and event years within documents, without assessing their assignment to individual events. GPT-5.4 ranked highest among ten LLMs (86.6\% F1). Verbatim evidence occurrence was 72.0\% for GPT-5 and 47.2\% for GPT-5.4, measuring textual traceability without establishing attribute support. We report production failure modes and automated label and location-rule compliance checks. Prompts, schema, and outputs will be released for adaptation to national reporting archives.

%\footnote{Our code and data 
%\ifarxiv
%are at \url{https://github.com/PreventionWeb/groundsource.git}. \textcolor{red}{maybe I should change the name of the repo so that it's not taking from google's \kem{no code needed at submission, remove this for anonymity}}
%\else
%have been uploaded to the submission system, and will be open-sourced upon acceptance.
%\fi
%}
\end{abstract}
\section{Introduction}
\label{sec:introduction}
Gaps in data availability, timeliness, and granularity constrain humanitarian analysis and development monitoring \citep{ocha2026humanitariandata,oecd2025sdgdatagaps}. Where structured records are scarce but narrative reports are available, information extraction offers a means of quickly obtaining source-linked records. The United Nations Office for Disaster Risk Reduction (UNDRR) maintains an extensive collection of disaster-risk-reduction documents through PreventionWeb, including reports, policy documents, assessments, and news articles.\footnote{\url{https://www.preventionweb.net/about-preventionweb}} Extracting structured hazard events from these narratives requires identifying reported occurrences and associating their hazard types, dates, and locations. 

Compared with event-focused news, narrative disaster-risk-reduction (DRR) documents introduce additional ambiguity. Reports can interleave policy recommendations, hypothetical scenarios, and retrospective accounts of multiple disasters. Dates may denote publication or policy periods. Locations may describe institutional jurisdictions rather than affected areas. Extraction therefore requires evidence linking each attribute to a specific occurrence.
Geoparsing methods combine place-name detection with contextual geocoding \citep{cafferata2026geolocation} or resolve existing disaster location descriptions to geometries \citep{ronco2025subnational}. Resolving a place name, however, does not establish which event affected that location.

\begin{figure}[H]
    \centering
    \includegraphics[width=1\linewidth]{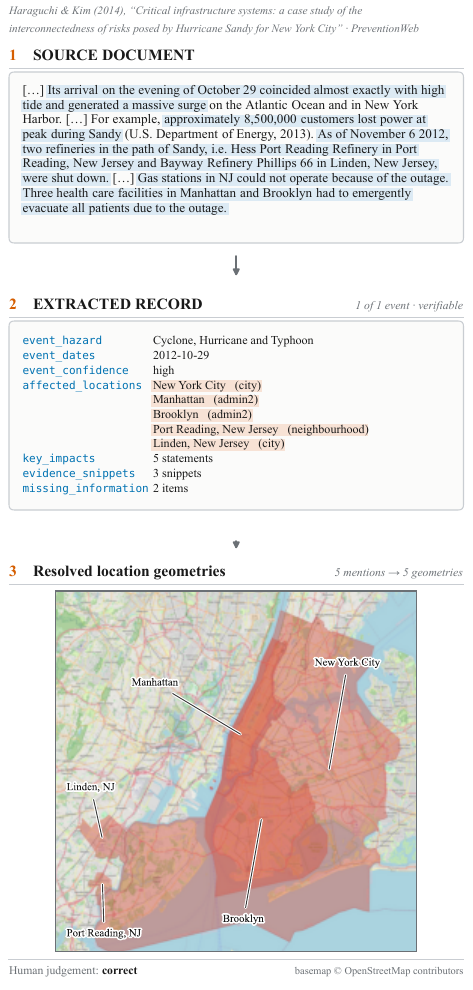}
    \caption{One document end to end. (1) An excerpt of the 12{,}000-character window passed to the model, with the spans returned in \texttt{evidence\_snippets} shaded. (2) The extracted record, abridged to counts for \texttt{key\_impacts}, \texttt{evidence\_snippets} and \texttt{missing\_information}. (3) The five location strings as the union of their resolved geometries. Judged correct (\S\ref{sec:spatial-output}). Basemap \copyright{} OpenStreetMap contributors.}
    \label{fig:example}
\end{figure}

This paper presents a large language model (LLM) pipeline combining document screening, event identification, argument extraction, and geocoding. Source evidence is retained for inspection and subsequent record consolidation. PreventionWeb provides the initial application corpus, with adaptation to national reporting archives as a design objective. Figure~\ref{fig:example} illustrates the workflow. Three contributions are presented:
\begin{itemize}
    \item \textbf{An auditable extraction framework.}
    Event records are defined through explicit evidence requirements, a controlled hazard vocabulary, and a configurable output schema.
    \item \textbf{A controlled empirical evaluation.}
    Extraction is evaluated on $217$ manually annotated documents against a spaCy--gazetteer baseline, with ten LLMs compared under the same schema, hazard vocabulary, document window, and geocoding procedure. We report document-level attribute recovery and limits of this evaluation for multi-event documents. 
    \item \textbf{Analysis of a corpus-scale batch run.}
    We characterise extraction yield, diagnostic failures, and geographical resolution across 10{,}000 documents, including a manual audit of 75 resolved location mentions.
\end{itemize}

\section{Background and Related Work}
\label{sec:background}

Document-level approaches link event arguments across sentences \citep{ebner-etal-2020-multi} and formulate argument extraction as conditional generation \citep{li-etal-2021-document}. HumSet supports humanitarian classification and span extraction \citep{fekih-etal-2022-humset}. Surveys cover crisis social media \citep{lewis2025tapping} and LLM applications in disaster management \citep{lei2025harnessing}. Explicit guidelines support instruction-based extraction \citep{sainz2024gollie,jiao-etal-2023-instruct}. Grammar-constrained decoding enforces output structure \citep{geng-etal-2023-grammar}. 

Geoparsing separates place-name detection from geographic resolution \citep{zhang-bethard-2023-improving}. \citet{cafferata2026geolocation} combine few-shot LLM-based NER with contextual geocoding and assess geographic and socioeconomic performance variation. \citet{ronco2025subnational} resolve EM-DAT location descriptions through LLM processing and cross-checks across GADM, OpenStreetMap, and Wikidata, with reliability scored by source agreement. 

Reference resources include EM-DAT event and impact records \citep{delforge2025emdat} and GDIS subnational locations \citep{rosvold2021gdis}. Comparisons of EM-DAT and DesInventar motivate accounting for event definitions and reporting coverage during evaluation \citep{panwar2020disaster}. Groundsource uses Gemini to extract approximately 2.6 million flood records with dates and locations from news \citep{mayo2026groundsource}. \citet{ronco2026storylines} use known EM-DAT events to retrieve Europe Media Monitor reports and generate storylines and knowledge graphs. \citet{decostanzi2026verifiable} enrich EM-DAT events with ReliefWeb and news documents, attach field-level citations, and evaluate citation quality. Like Groundsource, our pipeline identifies events without seed records. It extends verification-oriented extraction to multiple hazards in policy reports, assessments, and case studies. 

\section{Task and Data}
\label{sec:corpus-events}

\subsection{Corpus Description}
The system processes documents from PreventionWeb, the disaster-risk-reduction knowledge platform maintained by UNDRR. The collection includes heterogeneous materials, such as national risk assessments, technical reports, policy documents, containing retrospective disaster case studies.\footnote{\url{https://www.preventionweb.net/knowledge-base/type-content/documents-publications}}
The corpus, a collection exceeding 80{,}000 instances, was restricted to production corpus of 10{,}000 documents. The latter was predominantly English: 9{,}891 documents (98.9\%), compared with 52 Spanish, 27 Arabic, 16 Japanese, and 14 French documents. This distribution does not support reliable comparisons of extraction quality across languages. Catalogue entries provide descriptive metadata, including titles, source organisations, publication years, languages, and geographical tags as summarised in Table ~\ref{tab:metadata} for the retrieved documents. These fields describe source documents and are not extracted event attributes. Table ~\ref{tab:content-yield} also reports the counts of each content type within the selected corpus. 

\subsection{Event definition}
Extraction targets three attributes defining a specific hazard occurrence: hazard type, event timing, and directly affected locations. Hazard labels follow the PreventionWeb vocabulary, whose navigation categories appear in Table~\ref{tab:hazards}. Compound events receive combined labels ordered by prominence in the source, such as \texttt{Earthquake|Tsunami}. The generic label \texttt{Multi-hazard} is excluded; candidates without an applicable exact label are discarded.
Event dates preserve the finest temporal resolution supported by the source, including calendar dates, years, or short seasonal expressions. Publication, workshop, and report dates are excluded. Locations must identify directly affected places at the finest supported geographical granularity.

The model also extracts reported impacts, such as casualties and financial losses, evidence snippets, an ordinal confidence rating, and missing-information notes where applicable. Impacts require explicit source support; their absence does not invalidate an identifiable event. Confidence reflects the model's assessment of evidential support, not a calibrated probability of correctness \citep{pawitan-holmes-2025-confidence}.

Two Boolean fields flag event verifiability and document focus. Verifiability requires a supported hazard, at least one event date, at least one qualifying affected location, and \texttt{high} or \texttt{medium} confidence. The focus flag requires the document, or a qualifying bounded section, to centre substantively on the event; a valid reference within a broader policy or multi-event document may not qualify. Here, \emph{verifiable} means sufficiently documented for source-based checking, not independently confirmed.

\section{System}
\label{sec:system}

\subsection{LLM-Based Extraction}
The system applies the task definition through schema-constrained LLM extraction, eligibility filtering, and geocoding. Five prompt configurations encode different candidate-search and verification instructions (Table~\ref{tab:prompts}). Their evaluation is described in Section~\ref{sec:evaluations}.
The production run used GPT-5. Model comparisons, including GPT-5, were run separately through OpenRouter, which provides access to multiple model providers through a common API; validation scores are not computed from production outputs. 

Structured responses are specified using Pydantic models and generated through OpenAI Structured Outputs. Schema-constrained decoding restricts generation to the supported output structure, and the returned response is parsed into the corresponding typed object.
Table~\ref{tab:event-schema} specifies the event-level output contract. Structural compliance does not establish whether an extracted statement is supported by the document.

\paragraph{Geocoding and Spatial Mapping.}
Extracted location strings are subsequently resolved using Nominatim, an OpenStreetMap-based geocoding system. Textual queries return coordinates and, where available and requested, the geometry of the matched geographical object. The resulting linkage supports disaster risk communication, interactive visualisation, and spatial joins with external datasets. Their interpretation remains tied to the spatial resolution of the source: a point represents a resolved location, while an administrative polygon represents the corresponding territorial unit. Neither representation alone establishes the physical footprint of the hazard. 

\subsection{NER baseline.}
A rule-based baseline combined spaCy named entity recognition (\texttt{en\_core\_web\_lg}) with gazetteer lookup. The baseline and LLM processed the same documents, truncated to 12{,}000 characters, and used the same Nominatim geocoder. The baseline implemented the LLM's hazard vocabulary and normalisation rules as regular expressions. Generic terms such as \emph{risk}, \emph{hazard}, \emph{disaster}, and \emph{emergency} were excluded as triggers to limit extraction from abstract discussions.
Location candidates were obtained from \texttt{GPE}, \texttt{LOC}, and \texttt{FAC} entities. Temporal candidates were obtained from \texttt{DATE} entities. 

Each hazard mention defined a context window containing its sentence and the immediately preceding and following sentences. A candidate required at least one location and one normalisable date within this window. Candidates were grouped by their window's hazard-label set. Co-occurring hazards were represented by compound labels. Each record was limited to ten locations, five dates, five impacts, and three evidence snippets. Unsupported fields were left unfilled. Evidence snippets contained matched context sentences and four diagnostic flags recorded missing or weak extraction evidence.

\subsection{Production Execution}
Production execution was configured with a maximum of 10{,}000 input documents and a limit of 12{,}000 characters of document text per extraction input. Case studies, tables, or annexes can contribute to extraction only when their text is included within that limit, irrespective of the prompt's intended search scope. The document selection strategy depends on the order in which files were ingested into the source document store and this selection order is held fixed. Reported yields and composition describe this subset and need not represent the wider collection. Statistical use additionally requires event deduplication, compatible spatial units, and consideration of variation in reporting coverage.

\section{Evaluations}
\label{sec:evaluations}
\paragraph{Prompt Comparisons.}
Five variants for the extraction prompt were tested. These alternatives shared the output schema (see Tab ~\ref{tab:event-schema}), hazard vocabulary, and 12{,}000-character document window, differing only in their written instructions. The five prompt configurations were compared on a common set of 400 documents using GPT-5 with medium reasoning effort. Sampling used random seed (42). Further detail on the proposed prompts' characteristics and their comparison can be found in Appendix \ref{app:prompts}.
\paragraph{Evaluation.}
\label{sec:validation}
Human annotations of 217 document windows served as the evaluation ground truth. Annotation covered all qualifying events in each displayed window, and model outputs were not shown during annotation. The reference set comprised 217 document windows obtained through stratified sampling (Appendix~\ref{app:annotation-guidelines}). Detection was evaluated on all 217 windows, with 171 labelled \texttt{yes} and 46 treated as negative, comprising 42 \texttt{no} and four \texttt{unclear} annotations. 

Field extraction was evaluated only on the 171 positive windows, after pooling and deduplicating hazards, locations, and dates within each document. Consequently, field precision excludes predictions from reference-negative windows. All ten LLMs, including GPT-5, were evaluated through OpenRouter using a common extraction prompt, target schema, hazard vocabulary, and 12{,}000-character document-text limit. Complete-match accuracy measures whether all three document-level attribute sets match under the scoring rules. Neither pooled F1 nor complete-match accuracy evaluates event--argument assignment, including incorrect merging or splitting of occurrences.

\begin{table*}
\centering
\small
\setlength{\tabcolsep}{4pt}
\begin{tabular}{lrrrrr@{\hskip 12pt}rrrr@{\hskip 12pt}r}
\hline
& & \multicolumn{4}{c}{Pooled, all fields}
& \multicolumn{4}{c}{Per-field F1}
& \multicolumn{1}{c}{Evidence} \\
\cline{3-6}\cline{7-10}\cline{11-11}
Model & Docs & Acc. & P & R & F1
& Haz. & Loc-ex & Loc-fz & Date & Found (\%) \\
\hline
GPT-5.4 & 171 & 38.0 & \textbf{86.5} & 86.7 & \textbf{86.6} & 91.9 & 64.3 & \textbf{82.9} & 89.6 & 47.2 \\
GPT-5$^\dagger $ & 171 & \textbf{44.4} & 84.2 & 87.9 & 86.0 & \textbf{92.2} & \textbf{66.3} & 80.3 & \textbf{91.4} & 72.0 \\
Grok 4.6 & 171 & 32.2 & 75.0 & \textbf{88.6} & 81.2 & 89.0 & 58.8 & 74.6 & 88.2 & 46.0 \\
Claude Opus 5 & 171 & 31.6 & 78.9 & 83.6 & 81.2 & 89.8 & 59.1 & 76.5 & 84.0 & 85.6 \\
Claude Sonnet 5 & 171 & 37.4 & 79.0 & 83.0 & 80.9 & 89.7 & 58.1 & 75.8 & 84.3 & 90.8 \\
DeepSeek V3.1 & 171 & 31.0 & 78.4 & 81.0 & 79.7 & 86.8 & 61.6 & 73.7 & 85.1 & \textbf{94.6} \\
GPT-5 mini & 171 & 33.3 & 74.8 & 83.0 & 78.7 & 89.1 & 30.0 & 69.8 & 87.2 & 85.0 \\
Claude Haiku 4.5 & 171 & 31.6 & 83.5 & 73.8 & 78.3 & 85.0 & 59.5 & 74.5 & 80.8 & 51.5 \\
Llama 4 Maverick & 171 & 13.5 & 58.6 & 81.7 & 68.2 & 79.3 & 43.2 & 57.0 & 81.5 & 92.3 \\
Mistral Large & 171 & 8.2 & 53.7 & 60.5 & 56.9 & 73.4 & 37.2 & 47.0 & 65.9 & 83.8 \\
NER baseline & 171 & 1.2 & 35.9 & 57.6 & 44.2 & 67.4 & 20.1 & 30.3 & 57.5 & 100.0 \\
\hline
\end{tabular}
\caption{Document-pooled extraction scores over the 171 event-bearing documents of the 217-document validation set. \emph{Docs} reports the number of positive reference documents included in field evaluation. Complete-match accuracy uses these 171 documents as its denominator; micro-precision and micro-recall use predicted and reference attribute-item counts, respectively. The pooled columns count hazard matches, normalised fuzzy location matches, and year-set date matches over 1{,}416 reference items (263 hazard, 708 location, 445 date). Acc.: percentage of documents with no false positives or false negatives across all three fields; P: micro-precision; R: micro-recall; F1: micro-F1. The per-field columns give micro-F1 for each component separately under its own matching rule --- Haz.: hazard match; Loc-ex: exact location match after casefolding; Loc-fz: fuzzy match at token-set ratio $\geq$ 90 on normalised strings; Date: year-set match, likewise the pooled rule; Evidence found: percentage of model-quoted evidence snippets located in the source document; the NER baseline is 100\% by construction (control). All scores are percentages. Bold indicates the highest value per column. \\
$^\dagger$ Model and prompt used in the production run (\S\ref{sec:production-run}), re-run through OpenRouter for this comparison.}
\label{tab:validation-pooled}
\end{table*}

\section{Results}

\subsection{Prompt selection}
\label{sec:prompt-selection}
Prompt~05 yielded 58 high-confidence documents and 153 resolved geometry rows, compared with a maximum of 44 documents and 70 geometry rows among the other configurations. These observations supported selecting prompt~05 for production on the basis of extraction and geocoding yield. The selected prompt is reproduced in Appendix~\ref{app:prompt05}.
\subsection{Validation Results.}
The production configuration, GPT-5, reached a pooled F1 of 86.0 (P 84.2, R 87.9) and the highest complete-match rate (44.4\%), 41.8 points above the spaCy–gazetteer baseline in pooled F1 (Table~\ref{tab:validation-pooled}). On the full 217-window detection set, GPT-5 produced 165 true positives, 11 false positives, six false negatives, and 35 true negatives, corresponding to precision of 93.8\%, recall of 96.5\%, and F1 of 95.1\%. It returned a positive detection for 11 of the 46 windows treated as reference negatives (23.9\%), compared with 19 of 46 (41.3\%) for the baseline (Table~\ref{tab:validation-counts}; Appendix~\ref{app:valids}). It also achieved the highest F1 for hazard families (92.2), exact location matches (66.3) and event years (91.4). GPT-5.4 achieved the highest pooled F1 (86.6), precision (86.5) and fuzzy-location F1 (82.9), exceeding GPT-5 by 0.6 and 2.6 points respectively. 

Table~\ref{tab:validation-pooled} also reports the proportion of evidence snippets found verbatim in the searched source text. DeepSeek V3.1 had the highest rate among the LLMs (94.6\%); the baseline's 100\% rate follows from copying context sentences. This metric measures textual occurrence, not whether a snippet supports the attributes attached to it. Because the output contract permits paraphrases, failure to match verbatim does not by itself establish unsupported generation.

\section{Production Run and Analysis}
\label{sec:production-run}
\subsection{Extraction yield and corpus composition}
\label{sec:production-yield}

The run processed 10{,}000 documents and retained 3{,}572 event records satisfying the system's verifiability criteria. These records originated from 1{,}913 documents, corresponding to an extraction rate of 19.1\% (Table~\ref{tab:production-counts}). 
Among documents yielding at least one event, the mean was 1.87 records per document and the maximum was 40. Of these documents, 1{,}187 yielded one record, 385 yielded two, 158 yielded three, 71 yielded four, 101 yielded between five and ten, and 11 yielded more than ten. These are extraction counts and not event counts as separate documents may describe the same occurrence.
Extraction yield varied across source content types (Table~\ref{tab:content-yield}). Publications accounted for 67.6\% of processed documents and 1,580 of the 1,913 documents yielding events. Meeting and conference pages, represented by the source content type event, had a lower extraction rate of 8.9\%.

\begin{table}[t]
\centering
\small
\begin{tabular}{@{}lr@{}}
\toprule
\textbf{Measure} & \textbf{Count} \\
\midrule
Documents processed & 10{,}000 \\
Documents with at least one event & 1{,}913 \\
Documents without a retained event & 8{,}087 \\
Retained event records & 3{,}572 \\
High-confidence event records & 1{,}955 \\
Medium-confidence event records & 1{,}617 \\
\midrule
Location mentions & 5{,}463 \\
Resolved location mentions & 4{,}435 \\
Unresolved location mentions & 1{,}028 \\
Documents with at least one geometry & 1{,}686 \\
\bottomrule
\end{tabular}
\caption{Production counts at document, event-record, and location-mention levels. Event records have passed the operational extraction filter but have not necessarily been independently validated or deduplicated.}
\label{tab:production-counts}
\end{table}

\begin{table}[t]
\centering
\small
\begin{tabular}{@{}lrrr@{}}
\toprule
\textbf{Content type} &
\textbf{Documents} &
\textbf{With events} &
\textbf{Rate} \\
\midrule
Publication & 6{,}758 & 1{,}580 & 23.4\% \\
Meeting/conference & 2{,}086 & 185 & 8.9\% \\
News & 718 & 141 & 19.6\% \\
Vacancy & 431 & 7 & 1.6\% \\
Unknown & 7 & 0 & 0.0\% \\
\bottomrule
\end{tabular}
\caption{Documents yielding at least one retained event, grouped by upstream content type. Rates use the number of documents within each type as the denominator.}
\label{tab:content-yield}
\end{table}

Figure~\ref{fig:temporal-distribution} and Appendix~\ref{app:temporal} compare document publication years with extracted event years to characterise retrospective coverage. Publication years were available for 6{,}426 documents, with missingness primarily reflecting upstream metadata coverage. Event years were assigned to 3{,}485 records (97.6\%), with the largest annual counts in 2011 (322), 2010 (236), and 2004 (231), and none after 2021. Although source publication years begin in the 1980s, 76 event records refer to years before 1900. 

\subsection{Hazard composition and geometry linkage}
\label{sec:spatial-output}

The retained records comprised 1{,}955 high-confidence outputs (54.7\%) and 1{,}617 medium-confidence outputs (45.3\%). Fifty distinct hazard strings, including compound labels, were mapped to nine operational hazard families. The implementation's earthquake family, which includes tsunami labels, accounted for 1{,}227 records (34.4\%), followed by flood with 838 (23.5\%) and cyclone/storm with 384 (10.8\%). The remaining records were assigned to other hazards (358), drought (262), volcano (163), wildfire (162), landslide (101), and heat (77). Compound records were assigned to the first matching family; these counts therefore depend on the aggregation rule and do not enumerate every hazard component separately.

The 3{,}572 event records contained 5{,}463 location mentions, of which 4{,}435 were resolved to geometries, giving a mention-level resolution rate of 81.2\%. At least one geometry was obtained for 1{,}686 documents, equivalent to 16.9\% of the processed corpus and 88.1\% of documents yielding events. Among resolved mentions, 1{,}586 were classified at city level and 1{,}331 at first-order administrative level. Fallback queries were used in 171 cases, and 20 cases were flagged for potential geometry overshoot. 

Resolution success indicates that a geographical object was returned. It does not establish that the object denotes the place intended by the source or that its geometry represents the affected area. An audit assessed whether the returned object denoted the intended place and whether its geographical extent matched the object and granularity supported by the source. In a random sample of 75 of the 4,435 resolved mentions, 58 identified the correct place and extent (77.3\%; CI: 66.7–85.3\%), 16 identified the correct place but an incorrect extent (21.3\%; CI: 13.6–31.9\%), and one identified a different place (1.3\%; CI: 0.2–7.2\%). Thus, 74 resolutions identified the intended place (98.7\%; CI: 92.8–99.8\%), but this proportion includes geometries with incorrect extents. 

Automated checks and diagnostic inspection identified the failure categories in Table~\ref{tab:failure-taxonomy}; further details and examples appear in  Appendix~\ref{app:failures}. Vocabulary checks found no out-of-vocabulary hazard labels across 3{,}572 records containing 3{,}760 label components. None of the 200 country-only records violated the checked bare-country rule, while four records (0.1\%) contained country--subregion redundancy. These results establish compliance with the specific constraints checked. 

\section{Conclusion}
Constrained by a controlled hazard vocabulary, a fixed output schema and explicit evidence requirements, GPT-5 converted 10{,}000 PreventionWeb documents into 3{,}572 candidate georeferenced event records. On the 171 event-bearing documents of the reference set, the same model and prompt reached a document-pooled F1 of 86.0, against 44.2 for the spaCy–gazetteer baseline. These scores measure document-level attribute recovery, not the assignment of attributes to individual events. Future work will expand corpus coverage, normalise temporal expressions while preserving source-supported precision, and merge records describing the same occurrence across documents. 

Deployment taught three practical lessons. First, the input cap can exclude event evidence elsewhere in the document. Second, GPT-5 achieved higher F1 for hazard families than for affected-location strings. Third, the geometry audit identified extent mismatches even when the intended place was correctly resolved. The configurable prompt, vocabulary, and schema provide a basis for testing the pipeline on national reporting archives.

%\zhijing{Here is an example comment of mine.}
\section*{Limitations}
Institutional reports are long (median 30{,}040 characters) and describe events in an incidental manner. Because of this, any fixed-budget extractor model can only see parts of most documents. We capped inputs at 12{,}000 characters to keep the 10{,}000-document run affordable, instructed the model to search case studies and annexes within that window, and interpret ``no event extracted" as not assessed rather the document being event free. This cap is a parameter that can be raised if resources allow. 

PreventionWeb is predominantly English and draw from a single platform. Transferring this pipeline to national archives is ripe for future work.  Human annotations, especially for event-attribute linkage in multi-event documents are costly, so our reference set scores pooled attributes per document. We report per-field false positives and negatives so that error types are visible and explicitly caveat that event-level alignment remains an open issue.

The reference set over-represents event-bearing documents by design, so detection counts do not reflect corpus prevalence and field scores are conditional on documents containing events. The extraction prompt was selected using GPT-5 outputs, which may favour GPT-5 and models that behave similarly; differences under one F1 point among the top models should not be read as a ranking.

Regarding geocoding, any geocoder would return a best-matching object, and geometries do not represent a hazard footprint. We mitigate this by separating text-level location extraction from resolution, retaining admin-level tags and the original strings, and flagging fallback and overshoot cases for manual review. 

The outputs are candidate extractions requiring source-based verification. Retained evidence snippets and source links facilitate review but do not establish the correctness of each attribute or event association. Repeated accounts of the same occurrence may inflate record counts; cross-document consolidation remains necessary before interpreting them as counts of distinct events.

Our production pipeline uses proprietary models via a third-party API. While exact reruns cannot be guaranteed, code and derived outputs will be released upon publication to support reproducibility.

\section*{Ethical Considerations}
All inputs are publicly available PreventionWeb documents. We will release derived records, snippets, and source links, and use OpenStreetMap data under ODbL (© OpenStreetMap contributors). The pipeline extracts aggregate impacts and place names only. Records include confidence flags and provenance and are intended for human-reviewed enrichment of disaster databases only, not for any operational, insurance, or legal use. All annotations were performed by authors on the project.

\ifarxiv
%\section*{Author Contributions}\label{sec:contributions}

\section*{Acknowledgment}
This work was funded by the United Nations Office for Disaster Risk Reduction (UNDRR) and deployed utilizing their internal tools. The views expressed herein are solely those of the authors and do not necessarily represent the official position of the UNDRR.

\fi

\iffalse
\section{End of Main Paper}
\fi

\bibliography{refs}

\clearpage

\appendix

% This is an appendix.

\onecolumn

\section{Appendix A: Prompt for LLM-Based Event Extraction}
\label{app:prompt05}

\begin{tcblisting}{
    prompt,
    breakable,
    listing only,
    title={\textbf{LLM-extraction instructions: prompt 5}},
    listing options={
        language={},
        basicstyle=\ttfamily\scriptsize,
        columns=fullflexible,
        keepspaces=true,
        breaklines=true,
        breakatwhitespace=false,
        showstringspaces=false,
        tabsize=2,
        literate={—}{{---}}1
    }
}
==============================================================================
MESSAGE 1 — role: "system"
==============================================================================
You are a disaster-event verification analyst tuned for comprehensive extraction.

Use prompt 3 / prompt 4 style verification discipline as the backbone: every emitted
field must be supported by the source. Use prompt 1 / prompt 2 style recall only as a
search strategy: search widely across the document, but do not lower the evidence bar.

Search scope:
- Read the title, URL, metadata, and document text.
- Look for concrete past or current hazard events in the main topic, case studies,
  annexes, table narratives, country profiles, lessons learned, historical examples,
  background sections, dataset descriptions, and response or loss records.
- Return one event object per distinct concrete occurrence. Do not choose only one
  winner when multiple reliable events are supported.
- Reject forecasts, simulations, exercises, preparedness activities, meetings,
  methodologies, tools, generic risk descriptions, and future scenarios unless they also
  contain a concrete historical or current event with enough support for verification.

Evidence discipline:
- Use metadata, title, URL, countries, hazards, publication year, and common context only
  as hints. They cannot prove that an event happened, supply a missing date, supply an
  affected location, justify country-level scope, or fill in impacts.
- Omit passing mentions unless the same source gives explicit hazard, event timing,
  directly affected location, and high or medium confidence support.
- key_impacts and evidence_snippets must be grounded in the source. Do not invent
  numbers, dates, locations, or impacts.

Hazard classification:
For each returned event, event_hazard MUST contain only exact PreventionWeb hazard labels
from the list below.
Do not output synonyms, older local labels, broad custom labels such as Multi-hazard, or
source-specific wording.

Main PreventionWeb hazard labels:
Avalanche; Cold Wave; Cyclone, Hurricane and Typhoon; Drought and Desertification;
Earthquake; Epidemic and pandemic; Flood; Heatwave and Extreme Heat; Insect infestation;
Land subsidence; Landslide; Nuclear, biological, chemical (NBC); Sea level rise;
Technological hazard; Thunderstorm; Tornado; Tsunami; Volcano; Wildfire.

Other PreventionWeb hazard collection labels, allowed only when directly supported:
Sand and dust storm; Fall armyworm; Stampede and crowd collapse;
Geomagnetic storm and space weather; Human-induced earthquakes.

Normalize common source terms to the closest exact PreventionWeb label:
- hurricane, typhoon, tropical cyclone, tropical storm, tropical depression, storm surge
  -> Cyclone, Hurricane and Typhoon
- volcanic eruption, ash fall, lava flow, lahar -> Volcano
- disease outbreak, epidemic, pandemic -> Epidemic and pandemic
- heatwave, extreme heat, heat stress -> Heatwave and Extreme Heat
- locust outbreak, pest infestation, swarm -> Insect infestation
- chemical, nuclear, biological, radiological, contamination, gas leak, NaTech
  -> Nuclear, biological, chemical (NBC)
- explosion, collapse, dam failure, bridge failure, rail accident, transport accident,
  water supply failure, ICT outage, malware, urban fire -> Technological hazard
- mudslide, mud flow, debris flow, rockfall, lahar when described as a slope/mass
  movement -> Landslide
- flash flood, coastal flood, Glacial Lake Outburst Flood, snowmelt flood, fluvial flood,
  surface water flooding -> Flood

If multiple PreventionWeb labels are explicitly required for one event, join exact labels
with " | " in order of source prominence, e.g. "Earthquake | Tsunami". Every component
must be one of the exact labels above.

If the source describes a real event but no exact PreventionWeb label fits, do not return
that event in the events list. Do not infer beyond the source.

Date rules:
- event_dates must describe the hazard event itself, not publication, meeting, workshop,
  training, data-collection, or report-writing dates.
- Use the most precise source-supported form: YYYY-MM-DD, YYYY-MM, YYYY, or a short
  event-timing phrase when the timing is clear but not normalisable.
- If event timing is not source-supported, do not return the event.

Location rules:
- affected_locations must be places directly affected by this event: flooded, damaged,
  hit, evacuated, cut off, contaminated at the incident site, or otherwise directly
  impacted.
- location_admin_levels must be parallel to affected_locations, same order and same
  length, using only the allowed schema values.
- Do not emit a bare country for sub-national hazards. This includes Avalanche; Cyclone,
  Hurricane and Typhoon when the footprint is a landfall, storm track, or storm surge;
  Earthquake; Flood when the footprint is a flash flood, coastal flood, Glacial Lake
  Outburst Flood, basin, or named flooded area; Land subsidence; Landslide; Nuclear,
  biological, chemical (NBC); Technological hazard; Tornado; Tsunami; Volcano; Wildfire;
  plus oil spill, mudslide, dam burst, and transport accident.
- Emit a bare country only for Drought and Desertification; Heatwave and Extreme Heat;
  Cold Wave; Epidemic and pandemic; or Insect infestation when the text explicitly says
  the scope was nationwide and no sub-national affected location is named.
- For Technological hazard, Nuclear, biological, chemical (NBC), oil-spill, dam-burst,
  industrial, radiological, transport, or similar accidents, emit only the incident
  facility, named exclusion zone, or directly impacted settlements. Exclude downwind,
  plume, fall-out, or receiving countries and regions.
- When a country and specific affected sub-regions are both named for the same event,
  emit only the sub-regions.
- If the most granular location is too specific to map on its own, such as a room, pier,
  gate, hangar, vehicle, warehouse, building section, or sub-asset, also include its
  parent city or district as a separate item. Do not concatenate them.
- Geological features, named coasts, basins, fault lines, and large oceanic features are
  allowed when they are the most specific directly affected location stated by the source.

Admin-level tags:
- For each emitted location, emit one tag from: point, neighbourhood, city, admin2,
  admin1, country, coastal_zone, basin, feature, unknown.
- Use country only when the country-level rule above allows it.
- Use unknown only as a last resort.

Confidence and verifiability:
- high requires explicit hazard, valid affected location, event timing, and at least one
  impact or evidence item.
- medium requires explicit hazard and valid affected location, with event timing present
  but date precision or impact detail partial.
- low covers thin evidence, passing mentions, or any candidate made weak by the
  location/date rules.
- Return only events where is_verifiable_event=true: event date or timing, valid directly
  affected location, matching admin level, explicit hazard, and high or medium
  confidence all hold.
- Omit low-confidence candidates and events made unverifiable by the location/date rules.

is_single_actual_event:
- Set true only when the whole document or a clearly bounded multi-sentence section is
  substantively about that specific event.
- Otherwise set false, even when the event is valid and verifiable.

Return JSON matching the schema exactly, with a top-level events list. Return events=[]
when no verified reliable event satisfies these rules.

==============================================================================
MESSAGE 2 — role: "user"
==============================================================================
Search the full metadata and document text for every distinct concrete past or current disaster event, including events embedded in case studies, annexes, tables, lessons learned, examples, and background sections. Use high recall only for finding candidates; emit only source-supported high/medium-confidence events with hazard, event timing, valid directly affected locations, matching admin levels, and strict location/date discipline. Return JSON matching the schema exactly.

{
  "document_metadata": {
    "asset_key": "<asset_key>",
    "source_id": "<source_id>",
    "attachment_id": "<attachment_id>",
    "asset_url": "<asset_url>",
    "content_url": "<content_url>",
    "asset_type": "<asset_type>",
    "title": "<title>",
    "publication_year": "<publication_year>",
    "language": "<language>",
    "countries": "<countries>",
    "themes": "<themes>",
    "hazards": "<hazards>",
    "organizations": "<organizations>",
    "last_processed_at": "<last_processed_at>"
  },
  "document_text": "<document_text trimmed to --max-chars>"
}
\end{tcblisting}

\begingroup
\captionsetup{skip=3pt}
\captionof{figure}{%
System and user prompts for multi-hazard event extraction
(\texttt{prompt05}). Document text is truncated to
12{,}000 characters; placeholders are populated at runtime.
}
\label{fig:event-extraction-prompt}
\endgroup

\newpage

\section{Appendix B: PreventionWeb Hazard Labels \& Metadata}
\label{app:hazard-families}
\begin{table}[H]
\centering
\small
\begin{tabularx}{\columnwidth}{@{}>{\raggedright\arraybackslash}X@{}}
\toprule
\textbf{PreventionWeb hazard label} \\
\midrule
Avalanche \\
Cold Wave \\
Cyclone, Hurricane and Typhoon \\
Drought and Desertification \\
Earthquake \\
Epidemic and pandemic \\
Flood \\
Heatwave and Extreme Heat \\
Insect infestation \\
Land subsidence \\
Landslide \\
Nuclear, biological, chemical (NBC) \\
Sea level rise \\
Technological hazard \\
Thunderstorm \\
Tornado \\
Tsunami \\
Volcano \\
Wildfire \\
Sand and dust storm\\
Fall armyworm\\
Stampede and crowd collapse\\
Geomagnetic storm and space weather\\ 
Human-induced earthquakes.\\
\bottomrule
\end{tabularx}
\caption{Hazard labels displayed in PreventionWeb's website navigation. The extraction vocabulary preserves the exact labels fixed for the reported implementation.}
\label{tab:hazards}
\end{table} 

\begin{table}[H]
\centering
\small
\renewcommand{\arraystretch}{1.15}
\begin{tabular}{@{}p{0.21\textwidth}p{0.56\textwidth}p{0.18\textwidth}@{}}
\hline
\textbf{Metadata label} & \textbf{Description} & \textbf{Coverage} \\
\hline
\texttt{title}
& Title associated with the source document in PreventionWeb.
& 10{,}000 (100.0\%) \\
\texttt{publication\_year}
& Publication year of the document; distinct from the dates of events described in its text.
& 6{,}426 (64.3\%) \\
\texttt{language}
& Recorded language of the source document.
& 10{,}000 (100.0\%) \\
\texttt{countries}
& Country tags associated with the document; provide geographic context but do not establish an event's affected locations.
& 7{,}046 (70.5\%) \\
\texttt{themes}
& Thematic tags describing the document's disaster risk reduction topics.
& 9{,}271 (92.7\%) \\
\texttt{hazards}
& Hazard tags assigned to the document in PreventionWeb's taxonomy; their presence does not establish that an actual event is reported.
& 2{,}740 (27.4\%) \\
\texttt{organizations}
& Organizations associated with the document in its source metadata.
& 7{,}615 (76.2\%) \\
\texttt{content\_url}
& URL of the PreventionWeb content page associated with the document.
& 9{,}993 (99.9\%) \\
\hline
\end{tabular}
\caption{PreventionWeb descriptive metadata retained by the extraction pipeline. Coverage is the number and percentage of the 10{,}000 distinct document assets in the production output with a non-empty value, counting each asset once regardless of how many event records it produces. These figures describe metadata availability in the processed corpus. Country, theme and hazard tags describe documents not individual extracted events.}
\label{tab:metadata}
\end{table}

\newpage

\section{Appendix C: Schema-constrained LLM-Output}

\begin{table}[H]
\centering
\small
\setlength{\tabcolsep}{5pt}
\renewcommand{\arraystretch}{1.08}
\begin{tabularx}{\textwidth}{
@{}p{0.27\textwidth}
p{0.13\textwidth}
>{\raggedright\arraybackslash}X@{}}
\toprule
\textbf{Field} & \textbf{Type} & \textbf{Contract} \\
\midrule

\texttt{is\_single\_actual\_event} &
\texttt{bool}, required &
True only when the whole document, or a clearly bounded multi-sentence section, is substantively about the event. Otherwise false, including valid event references in policy or multi-event documents that lack such a qualifying section. \\
\addlinespace

\texttt{is\_verifiable\_event} &
\texttt{bool}, required &
True only when the hazard, at least one event date, and at least one rule-satisfying location are supported, with confidence assessed as high or medium. \\
\addlinespace

\texttt{event\_hazard} &
\texttt{str | null} &
Exact PreventionWeb labels only. Compound events join labels with \texttt{" | "} in order of source prominence. Synonymous output labels and \texttt{Multi-hazard} are prohibited. Candidates without an applicable exact label are discarded. \\
\addlinespace

\texttt{event\_dates} &
\texttt{list[str]} &
Event timing only; publication, workshop, and report dates are excluded. Preserve the most precise supported form: \texttt{YYYY-MM-DD}, \texttt{YYYY-MM}, \texttt{YYYY}, or a short expression such as ``monsoon season 2017''. \\
\addlinespace

\texttt{affected\_locations} &
\texttt{list[str]} &
Directly affected places, preferably granular. Sub-map-scale references, such as a room, pier, or vehicle, require a parent city or district as a separate list item, rather than concatenated text. \\
\addlinespace

\texttt{location\_admin\_levels} &
\texttt{list[str]} &
Parallel to \texttt{affected\_locations}, with identical length and ordering. One tag per location: \texttt{point}, \texttt{neighbourhood}, \texttt{city}, \texttt{admin2}, \texttt{admin1}, \texttt{country}, \texttt{coastal\_zone}, \texttt{basin}, \texttt{feature}, or \texttt{unknown}. \\
\addlinespace

\texttt{key\_impacts} &
\texttt{list[str]} &
At most five concise statements explicitly supported by the text. Qualitative descriptions are permitted when numerical information is absent. Numerical values must not be invented. \\
\addlinespace

\texttt{event\_confidence} &
\texttt{str | null} &
Expected values are \texttt{high}, \texttt{medium}, and \texttt{low}. The field is a free string rather than a Pydantic \texttt{Literal}; admissible values are constrained by the prompt and downstream filter. \\
\addlinespace

\texttt{evidence\_snippets} &
\texttt{list[str]} &
At most three short excerpts or paraphrases, each containing no more than 25 words. Extended quotations are excluded. \\
\addlinespace

\texttt{missing\_information} &
\texttt{list[str]} &
Free-text descriptions of evidential weaknesses or exclusions, such as an unavailable exact date or locations removed under an extraction rule. \\
\bottomrule
\end{tabularx}
\caption{Event-level output contract. Location tags are defined in Appendix~\ref{app:prompt05}. The contract combines structural requirements with semantic rules whose enforcement is not implied by the listed Python types.}
\label{tab:event-schema}
\end{table}
\newpage
\section{Appendix D: Prompt Comparison}
\label{app:prompts}

We compared five prompt configurations for multi-hazard event extraction on a common set of 400 documents. The configurations differed in the scope of candidate identification, the placement of verification instructions, and the emphasis on schema compliance (Table~\ref{tab:prompts}). The comparison assesses operational outputs from each configuration; it does not isolate the contribution of individual prompt instructions.

\begin{table}[H]
\centering
\small
\setlength{\tabcolsep}{5pt}
\begin{tabularx}{\textwidth}{
@{}p{0.10\textwidth}
p{0.34\textwidth}
>{\raggedright\arraybackslash}X@{}}
\toprule
\textbf{ID} & \textbf{Name} & \textbf{Design} \\
\midrule
\texttt{prompt01} &
\texttt{baseline\_strict\_location} &
Baseline schema with strict rules for identifying directly affected locations. \\
\addlinespace
\texttt{prompt02} &
\texttt{case\_study\_high\_recall} &
Broad candidate search across case studies, annexes, tables, and lessons-learned material. \\
\addlinespace
\texttt{prompt03} &
\texttt{verification\_first} &
Verification criteria are applied before event extraction. \\
\addlinespace
\texttt{prompt04} &
\texttt{compact\_schema\_focused} &
Minimal instructions centred on the required output schema. \\
\addlinespace
\texttt{prompt05} &
\texttt{unified\_verification\_recall} &
Broad candidate identification followed by a verification gate before emission. \\
\bottomrule
\end{tabularx}
\caption{Prompt configurations tested for multi-hazard event extraction. The descriptions indicate design objectives, rather than measured performance.}
\label{tab:prompts}
\end{table}

Table~\ref{tab:prompt-selection} reports the number of documents assigned high confidence by the model, the number of resolved geometry rows, and the numbers of assets marked as resolved or unresolved by the geocoding stage. High confidence is a model-assigned category, rather than an independently verified correctness label. Likewise, geocoding resolution indicates that a geographic result was obtained; it does not establish that the result identifies the intended place or its affected extent. Geometry-row counts also do not establish the number of distinct hazard events.

\begin{table}[H]
\centering
\small
\begin{tabular}{@{}lrrrr@{}}
\toprule
\textbf{Prompt} &
\textbf{High-confidence documents} &
\textbf{Resolved geometry rows} &
\textbf{Resolved assets} &
\textbf{Unresolved assets} \\
\midrule
01 & 37 & 66 & 31 & 6 \\
02 & 41 & 63 & 39 & 2 \\
03 & 44 & 70 & 39 & 5 \\
04 & 34 & 59 & 29 & 5 \\
\textbf{05} & \textbf{58} & \textbf{153} & \textbf{50} & \textbf{8} \\
\bottomrule
\end{tabular}
\caption{Operational outputs from the 400-document prompt comparison. High confidence is a model-assigned category; resolved assets and geometry rows measure whether the geocoding worked not its correctness.}
\label{tab:prompt-selection}
\end{table}

\texttt{prompt05} produced the largest numbers of high-confidence documents (58), resolved geometry rows (153), and resolved assets (50). Compared with the largest corresponding counts among the other configurations, these results indicate greater output volume under the combined candidate-identification and verification instructions.

\newpage
\section{Appendix E: Annotation Guidelines}
\label{app:annotation-guidelines}

\paragraph{Ground-truth construction.}
The reference set was built in two stages. First, the 10{,}000 processed documents were divided by production output into an event stratum of 1{,}913 documents yielding at least one retained event record and 8{,}087 documents yielding none. Second, using random seed 42, 167 documents were drawn uniformly from the event stratum (the event arm), and 50 were drawn uniformly from the 9{,}833 documents not already selected (the random arm). The random arm was therefore not restricted to documents without retained records. Of the 167 event-arm documents, 157 were labelled yes, as were 14 of the 50 random-arm documents, including all 11 production-flagged ones. The reference set comprised 215 English and 2 Japanese documents. Over-representing event-bearing documents was necessary to allow event-identification validation statistics. Production outputs determined sampling strata only; all models were rerun on the same sampled documents and evaluated against independently annotated human ground truth, without reusing production predictions for scoring. 

\paragraph{Annotation unit and task.}
The annotation unit was the fetched document-text prefix, capped at 12{,}000 characters. The annotator answered whether this window reported at least one specific, verifiable event, using \texttt{yes}, \texttt{no}, or \texttt{unclear}. Annotation instructions required all qualifying occurrences in the displayed window to be considered, regardless of the stratum from which the document was drawn. Model outputs were not shown during annotation. Hazards, dates, locations, and impacts were recorded as document-level lists. Hazards were only selectable from the PreventionWeb-label set. Dates, locations, and impacts were entered as pipe-separated values. An uncertainty checkbox and free-text notes were also available.
Of the 217 completed annotations, 171 were labelled \texttt{yes}, 42 \texttt{no}, and 4 \texttt{unclear}. Both \texttt{no} and \texttt{unclear} were coded as detection negatives. Field extraction was evaluated on the 171 positive documents. Table~\ref{tab:annotation-disposition} summarises these assignments.

\begin{table}[htbp]
\centering
\small
\begin{tabular}{@{}lrll@{}}
\toprule
Reference label & Documents & Detection & Field extraction \\
\midrule
\texttt{yes}     & 171 & Positive & Included \\
\texttt{no}      &   42 & Negative & Excluded \\
\texttt{unclear} &   4 & Negative & Excluded \\
\bottomrule
\end{tabular}
\caption{Reference-label disposition. Detection uses all 217 documents; field extraction uses the 171 documents labelled \texttt{yes}.}
\label{tab:annotation-disposition}
\end{table}

\newpage

\section{Appendix F: Failure Taxonomy \& Common Errors}
\label{app:failures}
\begin{table}[H]
\centering
\small
\setlength{\tabcolsep}{5pt}
\begin{tabularx}{\textwidth}{
@{}>{\raggedright\arraybackslash}p{0.155\textwidth}
>{\raggedright\arraybackslash}X
>{\raggedright\arraybackslash}p{0.30\textwidth}@{}}
\toprule
\textbf{Category} &
\textbf{Observed pattern and interpretation} &
\textbf{Example(s)} \\
\midrule

Temporal normalisation &
The first date is expressed as a free phrase, covering 428 distinct surface
forms. Intervals, seasons, decades, and event-relative expressions require
interpretation beyond direct ISO parsing. Such expressions are not necessarily
extraction errors. &
\texttt{2017 (meteorological year: May 2017 -- April 2018)} ($n{=}11$), a
parenthetical gloss written into a date field; also
\texttt{2005-08-25 to 2005-08-30}, \texttt{summer 2007}, \texttt{1970s},
\texttt{2017 Atlantic hurricane season}. \\
\addlinespace

Potential publication-date substitution &
All extracted dates have year precision and equal the document's publication
year. This is a candidate indicator of metadata leakage; contemporaneous
reporting can produce the same pattern legitimately. &
\texttt{7260abd2fda7c13c} --- Drought, \texttt{2018}, Cape Town, in a document
published in 2018. The ``Day Zero'' drought spans 2015--2018, so the year is
defensible but indistinguishable from a metadata default. \\
\addlinespace

Synthesised temporal anchors &
Relative expressions are combined with document-date fragments, for example,
``past two years before 2010-10''. A parser may incorrectly interpret the
embedded fragment as the event date. &
\texttt{3090f7dd140e14b7} --- \texttt{past two years before 2010-10} on six of
seven events, and \texttt{most recently before 2010-10-28} on the seventh,
where the evidence field itself records only ``in the past two years''. \\
\addlinespace

Geographical role confusion &
Locations were flagged as possible hazard origins or associated features rather
than directly affected places, including epicentres, faults, facilities, and
rivers. Resolving these mentions can produce geometries inconsistent with the
intended affected-location relation. &
\texttt{10fea7a69d549bb8} --- Earthquake, 1976, sole location
\texttt{Motagua Fault}; \texttt{b7a109e614530d13} --- Landslide,
\texttt{Hunza River}; \texttt{80d28a2163122619} ---
\texttt{Deep Horizon oil rig \textbar{} Louisiana's coastal waters}, all rejected by the human annotator. \\
\addlinespace

Descriptive location expressions &
Descriptions are emitted in place of sufficiently specified toponyms,
complicating candidate generation and geographical disambiguation. &
\texttt{Brisbane River catchment area}, \texttt{coastal areas of Nias},
\texttt{Texas--Louisiana border area}, \texttt{Colchester area}. Faithful to
the source and unresolvable by the gazetteer. \\
\addlinespace

Administrative redundancy &
A country is emitted alongside its own subregions, contrary to the applicable
output rule. This can duplicate spatial representations at different
administrative scales. &
\texttt{eefa9abdf1c50ae5} --- Earthquake, 1988,
\texttt{Spitak \textbar{} Armenia} tagged \texttt{city \textbar{} country}, so
the same event carries both a point-scale and a national footprint. \\
\bottomrule
\end{tabularx}
\caption{Diagnostic taxonomy for the production output. Examples are quoted verbatim from the production run and keyed by \texttt{asset\_key} prefix.}
\label{tab:failure-taxonomy}
\end{table}

\newpage
\section{Appendix G: Further Details on the LLM-Model Validation}
\label{app:valids}
\begin{table}[H]
\centering
\small
\setlength{\tabcolsep}{3.5pt}
\begin{tabular}{lrr@{\hskip 10pt}rrrr@{\hskip 10pt}rrr@{\hskip 8pt}rrr@{\hskip 8pt}rrr}
\hline
& \multicolumn{2}{c}{Documents}
& \multicolumn{4}{c}{Event detection}
& \multicolumn{3}{c}{Hazard family}
& \multicolumn{3}{c}{Locations}
& \multicolumn{3}{c}{Dates} \\
\cline{2-3}\cline{4-7}\cline{8-10}\cline{11-13}\cline{14-16}
Model & Ret. & Err. & TP & FP & FN & TN
& TP & FP & FN & TP & FP & FN & TP & FP & FN \\
\hline
GPT-5.4 & 217 & 0 & 164 & 11 & 7 & 35 & 245 & 25 & 18 & 597 & 136 & 111 & 386 & 31 & 59 \\
GPT-5 & 217 & 0 & 165 & 11 & 6 & 35 & 249 & 28 & 14 & 582 & 160 & 126 & 413 & 46 & 32 \\
Grok 4.6 & 217 & 0 & 160 & 13 & 11 & 33 & 247 & 45 & 16 & 604 & 307 & 104 & 404 & 67 & 41 \\
Claude Opus 5 & 217 & 0 & 159 & 11 & 12 & 35 & 243 & 35 & 20 & 578 & 226 & 130 & 363 & 56 & 82 \\
Claude Sonnet 5 & 217 & 0 & 160 & 11 & 11 & 35 & 239 & 31 & 24 & 565 & 218 & 143 & 371 & 64 & 74 \\
DeepSeek V3.1 & 217 & 0 & 162 & 11 & 9 & 35 & 239 & 49 & 24 & 534 & 207 & 174 & 374 & 60 & 71 \\
GPT-5 mini & 217 & 0 & 168 & 14 & 3 & 32 & 245 & 42 & 18 & 531 & 283 & 177 & 399 & 71 & 46 \\
Claude Haiku 4.5 & 217 & 0 & 149 & 9 & 22 & 37 & 210 & 21 & 53 & 509 & 150 & 199 & 326 & 36 & 119 \\
Llama 4 Maverick & 217 & 0 & 167 & 16 & 4 & 30 & 249 & 116 & 14 & 507 & 564 & 201 & 401 & 138 & 44 \\
Mistral Large & 217 & 0 & 135 & 14 & 36 & 32 & 199 & 80 & 64 & 388 & 556 & 320 & 270 & 104 & 175 \\
NER baseline & 217 & 0 & 153 & 19 & 18 & 27 & 213 & 156 & 50 & 306 & 1008 & 402 & 296 & 289 & 149 \\
\hline
\end{tabular}
\caption{Raw counts and confusion matrix behind Table~\ref{tab:validation-pooled}, over the 217 human-annotated documents. \emph{Ret.}: documents the run returned at least one non-error row for; \emph{Err.}: documents it returned only error rows for, excluded from that run's denominators.}
\label{tab:validation-counts}
\end{table}

\paragraph{Attribute matching and aggregation.}
Within each document, hazards, locations, and dates were deduplicated across occurrences. The primary field scores used hazard families, fuzzy matching of normalised location strings, and sets of event years. Location matching used greedy one-to-one assignment with RapidFuzz token-set similarity at a threshold of 90. Micro-F1 was computed from aggregated true-positive, false-positive, and false-negative counts. Pooled micro-F1 combined these counts across the three fields. This protocol evaluates document-level attribute recovery; event--argument assignment is outside its scope.

\paragraph{Run coverage.}
Eleven runs covered identical evaluation documents: ten OpenRouter model runs and the NER baseline. GPT-5 was evaluated in its own OpenRouter run under the production prompt, schema, and 12{,}000-character window; its scores are not computed from the stored production output. Each run covered 217 detection documents and 171 field-evaluation documents. Because event-bearing documents were over-represented by design (Appendix~\ref{app:annotation-guidelines}), field scores are conditional on documents that contain events, and detection counts reflect the sample's composition rather than the corpus.

\newpage

\section{Appendix H: Temporal Resolution \& Bias}
\label{app:temporal}

The temporal results show a concentration of recent publications without an equivalent concentration of recent event dates. Among the 6,426 documents with a parsable publication year, 73.1\% were published between 2010 and 2021. These documents account for 71.9\% of event records whose source publication year is known. However, only 44.8\% of the 3{,}485 event records with a parsable event year fall within that period. The publication timeline therefore describes when information entered the documentary record, while the event timeline reveals substantially broader historical coverage.

\begin{figure}[H]
    \centering
    \includegraphics[width=1\linewidth]{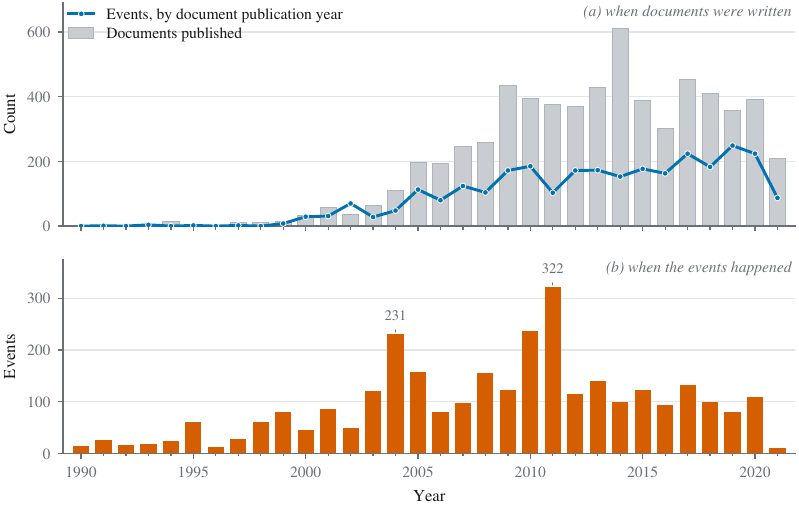}
    \caption{Temporal profile of the production run. (a) Documents by publication year (bars, $n=6{,}392$ of 10{,}000) and the extracted event records placed at their source document's publication year (line, $n=2{,}909$). (b) The corresponding records placed at their own event date ($n=3{,}046$). Counts are extracted records, not distinct events: cross-document deduplication is not applied. A record with several dates or a date interval is counted once, at its earliest date. Unparsable years and pre-1990 counts are omitted.}
    \label{fig:temporal-distribution}
\end{figure}

Historical coverage is particularly evident before 2000: just 67 dated documents were published in that period, whereas 778 dated event records—22.3\%—refer to events occurring before 2000, including 439 before 1990. Event counts do not rise steadily towards the most recent years. This pattern is consistent with the nature of the corpus: reports, assessments and guidance documents revisit earlier disasters as case studies and background.
Retrospective accounts in this institutional corpus therefore extend coverage beyond the publication period and demonstrate that extraction is not restricted to contemporaneous reporting. Historical coverage extends beyond the publication period, but the corpus may over-represent disasters repeatedly discussed for their impacts or policy significance. Moreover, publication years are unavailable for 35.7\% of documents, limiting conclusions about source coverage. Neither timeline should be interpreted as a measure of changes in physical disaster frequency: the counts reflect document availability, reporting choices and extraction, and may include repeated references to the same disaster.

\newpage
\end{document}